\documentclass[3p]{elsarticle}
\usepackage{graphicx}
\usepackage{multirow}
\begin{document}

\title{Population stratification for prediction of mortality in post-AKI patients}

\author[1,2]{Flavio S Correa da Silva\corref{ca}}\ead{fcs@usp.br}
\author[2]{Simon Sawhney}
\address[1]{Institute of Applied Health Sciences, University of Aberdeen AB25 2ZD United Kingdom}
\address[2]{Institute of Mathematics and Statistics, University of Sao Paulo 05508090 Brazil}
\cortext[ca]{Corresponding author.}
\begin{abstract} 
Acute kidney injury (AKI) is a serious clinical condition that affects up to 20\% of hospitalised patients. AKI is associated with short term unplanned hospital readmission and post-discharge mortality risk. Patient risk and healthcare expenditures can be minimised by followup planning grounded on predictive models and machine learning. Since AKI is multi-factorial, predictive models specialised in different categories of patients can increase accuracy of predictions. In the present article we present some results following this approach.
\end{abstract}

\maketitle
\section{Introduction}
\label{sec:intro}

Acute kidney injury (AKI) is a serious clinical condition that affects 10-20\% of hospitalised patients. AKI is associated with increases in (1) post-discharge mortality risk, (2) length of hospital stay and (3) healthcare expenditures \cite{thongprayoon2021clinically}, as well as short term unplanned re-admissions and mid term progressive chronic conditions. Around 33\% of AKI patients require unplanned re-admissions within 90 days after discharge and around 15\% develop progressive chronic kidney disease over the first year after discharge \cite{sawhney2017post,sawhney2021validation}.

AKI is multi-factorial, and accurate follow up planning is challenging. Machine learning has been viewed as promising to build tools to support decision making in clinical follow-up planning. Broadly speaking, recent initiatives can be structured along two alternatives:

\begin{enumerate}
    \item Tools grounded on prior medical expert knowledge, which is used to stratify patients according to meaningful attributes, in such way that specialised plans can be devised for each group of patients \cite{forte2021identifying,peng2021development,sawhney2018acute,thongprayoon2021clinically,zheng2021subtyping}.
    \item Tools grounded on machine learning techniques, which take control of the planning process and build accurate decision procedures which, however, demand extreme care in selection of new patients, to ensure compliance with population definitions that are used during preparation of decision procedures \cite{ashfaq2019readmission,choudhury2018evaluating,ebrahimzadeh2018prediction,hammoudeh2018predicting,wang2018predicting,xiao2018readmission}.
\end{enumerate}

Compliance with ethical standards demands that such tools are \textit{fair, transparent}, and \textit{optimised for the benefit of patients.} Technical requirements to ensure ethical compliance must include \textit{algorithmic transparency} to support fairness and transparency in decision making and \textit{optimised, goal-oriented patient stratification} to ensure human-centred optimised performance.

The research initiative presented in this article focused on the development of a tool to support clinical follow up planning for post-AKI patients after hospital discharge, with particular attention to ethical compliance based on technical requirements. We developed:

\begin{itemize}
    \item An automated procedure to stratify patients into groups in order to optimise prediction models for 90-days post-discharge mortality risk for each group, considering that the interpretations of descriptive attributes can vary across groups; and
    \item An automated procedure to allocate new patients into groups and corresponding prediction models, based on effective heuristics. 
\end{itemize} 

Patient stratification is inspired by the notion of \textit{Clustering Aware Classification} \cite{srivastava2023clustering}: we start with a stratification of patients in groups based on similarity clustering and considering a set of expert selected attributes to characterise individual patients, and then repeatedly reallocate patients across groups to increase accuracy of risk predictions. 

Accuracy of predictions is estimated, in our proposed method, based on the sum of \textit{Areas Under the Receiver Operating Characteristic} curve (\textit{AUROC}) \cite{flach2011coherent} across all groups, hence on how well, on average, the prediction models discriminate patients according to mortality risk. The selected stratification is the one that maximises the sum of \textit{AUROC} considering all groups. The choice of the metric to estimate accuracy is an important contribution in our work, targeting specifically the ethical issues related to (1) \textit{fairness}, (2) \textit{transparency}, and (3) \textit{patient-centred performance optimisation}, as it aims at resp. (1) \textit{the optimised characterisation of effectiveness of follow up planning considering all patient strata simultaneously}, (2) \textit{foundations of metrics on well stated statistical and computational methodology}, and (3) \textit{focus on a commonly accepted surrogate measure for wellbeing and quality of life} (namely, short-term mortality risk).

In order to assess the accuracy of predictions, a classification algorithm based on supervised learning is selected, and then used to discriminate patients based on predicted mortality risk. The selection of the algorithm is based on expert advice, as the presuppositions upon which the algorithm of choice is grounded (e.g. linear separability, or -- as in our case -- separability based on smooth differentiable manifolds as determined by generalised additive models) must reflect the epistemology of the domain.

The heuristic method employed to allocate new patients to groups must be aligned to the iterative process that generates the groups, i.e. it must be such that, if the new patients belonged to the initial population, they would most likely be allocated by the iterative process to the same groups to which they were allocated by the heuristic method. If the new patients are similar to the initial population (more technically: if the probability distribution in the population of new patients is the same as the probability distribution in the initial population), then the accuracy of predictions shall be preserved. Reliable estimates of accuracy are presented, based on sufficiently large samples of new patients.

Predictions for 90-days post-discharge mortality risk which are built following this process are grounded on several assumptions:

\begin{itemize}
    \item An initial population of patients is selected to build groups and corresponding risk prediction models. This population is \textit{assumed} to represent well the overall population of patients to be met in the future -- in the case of the present study, post-AKI patients after hospital discharge, for whom risk predictions must be built to support clinical follow up planning.
    \item A set of attributes is selected based upon expert advice, to portray the status of each patient in the initial population. This set of attributes is \textit{assumed} to characterise every relevant aspect of patients with respect to the health conditions under consideration.
    \item A family of prediction models is selected, also based upon expert advice, and \textit{assumed} to be epistemologically aligned to the structure of the domain under consideration.
    \item New patients are \textit{assumed} to be similar to those patients considered in the initial population, at least with respect to the extent to which the selected characterising attributes correlate with the predicted variable.
\end{itemize}

As a consequence, predictions are, by design, partially reliable, and should be considered only as support indicators for clinicians, to be taken under consideration together with other criteria. This is the reality of most machine learning techniques when applied to real life complex problems, despite not being so commonly presented explicitly as a fact.

In a sense, our work is similar to the initiative based on \textit{Deep Mixture Neural Networks} \cite{li2020predicting}. For the benefit of transparency, we have employed predictors based on \textit{Generalised Additive Models} (\textit{GAM}) \cite{hastie2017generalized} for predictions instead of deep neural networks, given that:

\begin{itemize}
    \item \textit{GAM}-based predictors are easier to understand intuitively and to analyse formally, and
    \item The sample complexity of \textit{GAM}-based predictors is orders of magnitude lower than that of deep neural networks, hence it is possible to build significantly more reliable predictors based on these predictors than on deep models given a fixed-sized sample \cite{srivastava2023clustering}.
\end{itemize}

In section \ref{sec:model} we present our proposed model for patient stratification in groups, allocation of new patients to existing groups, and assessment of risk prediction. In section \ref{sec:results} we present our experimental results and a brief discussion. Finally, in section \ref{sec:conclusion} we present our conclusions and proposed future work.

\section{Stratification based predictions}
\label{sec:model}

The fundamental premise on which we ground our model is that post-AKI patients can be grouped according to different profiles characterised by ranges of values of descriptive variables, and that an appropriate grouping procedure can optimise the accuracy of predictions about important future events -- e.g. mortality risk -- given that the interpretation of descriptive variables and what they can inform about future events can vary across groups. 

The utility of the identification of ranges of values that characterise groups of patients in such way that accuracy of predictions is optimised, however, depends on the ability to allocate new patients into appropriate groups, in such way that the optimised accuracy can be observed also in a data set of new patients.

\begin{figure}
    \centering
    \includegraphics[scale=0.45]{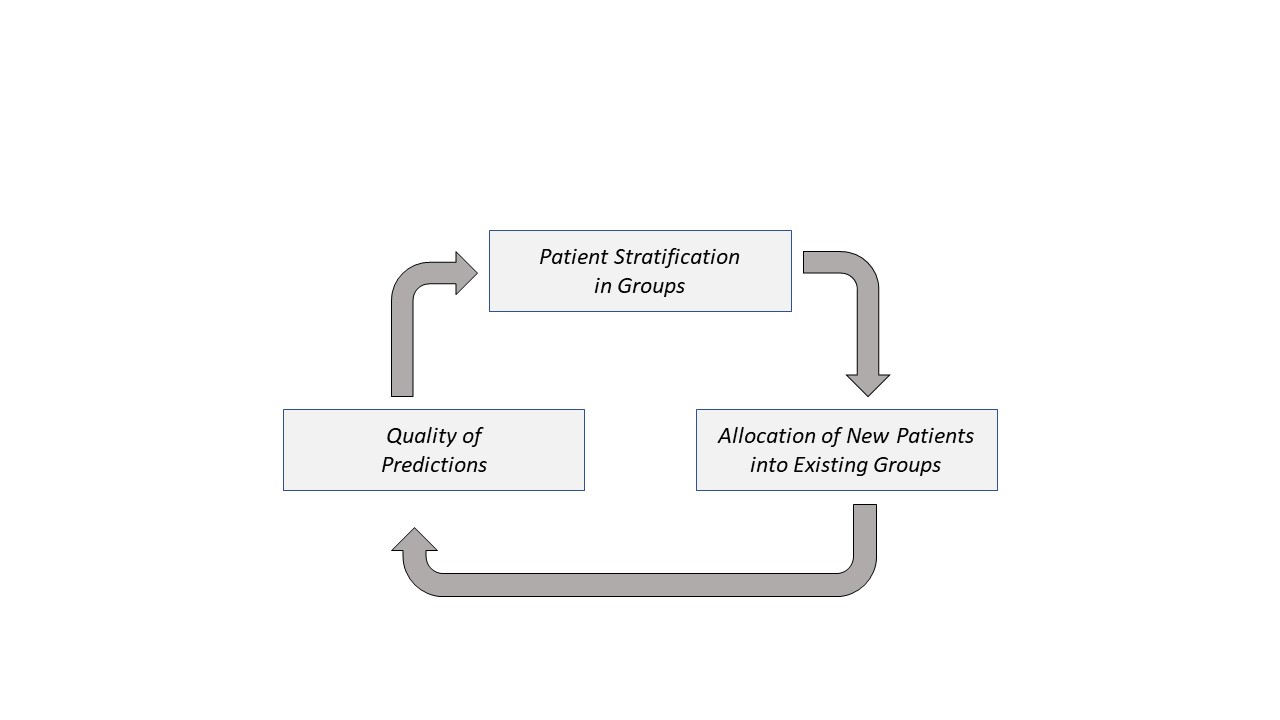}
    \caption{Circularity in optimised stratification of patients in groups}
    \label{fig:00}
\end{figure}

Instead of building groups based on clinical presuppositions about patient hallmarks, we adopt a method that allocates patients into groups in such way that predictions are explicitly and directly optimised, based on heuristics which are intuitively meaningful and can be empirically assessed. This problem is difficult given its inherent circularity (Figure \ref{fig:00}): stratification of patients in groups depends on the quality of predictions made for each group, allocation of new patients into existing groups depends on the definition of groups and quality assessment of predictions depends on the allocation of new patients into groups.

The solution we propose is inspired by the technique called \textit{Clustering Aware Classification} \cite{srivastava2023clustering} and is based on the idea of randomised hill-climbing given an uncharted search space for which parametric probabilistic assumptions are not available. We start with an arbitrary definition of patient groups using a \textit{training data set}, based on which a collection of predictors is built and applied to new patients belonging to a separate \textit{validation data set}, who are allocated to the groups according to a heuristic method, this way producing a heuristic evaluation of the quality of the groups as measured by the accuracy of their corresponding predictors. Then, random small perturbations are performed on the initial groups, by reallocating small chunks of patients in the \textit{training data set} across groups, after which patients in the \textit{validation data set} are reallocated to the newly created groups and the quality of the newly created predictors is assessed. If the new groups obtained after a perturbation present an improved heuristic evaluation of quality in predictions in comparison with the previous groups, then the new groups are preserved and the process is repeated; otherwise, the new groups are discarded and the process is repeated. This procedure is repeated until quality assessment reaches a point of stability.

This solution is implemented as follows:

\begin{enumerate} 

    \item A data set of patients is partitioned in three subsets, resp.:
    \begin{enumerate}
        \item A \textit{training data set}, used to build the \textit{reference groups} of patients.
        \item A \textit{validation data set}, used for heuristic assessment of patient groups.
        \item A \textit{test data set}, used to assess the expected final quality of classification based on groups.
    \end{enumerate}
    
    In our experiments, we have employed 50\% of the available data set for \textit{training}, 10\% for \textit{validation} and 40\% for \textit{testing}, out of an overall data set comprising $\approx 20k$ patients.
    
    \item A comprehensive collection of patient attributes is selected by an expert clinician. These attributes must include a decision labelling attribute to be used to build predictors based on supervised learning.
    
    In our experiments, we have adopted mortality prior to 90 days after patient discharge \textit{[Y/N]} as the decision labelling attribute. As for descriptive attributes, we have adopted the following:
    \begin{itemize}
        \item Patient gender
        \item Patient age at date of hospital admission
        \item Patient creatinine level taken at day of hospital discharge
        \item Patient maximum creatinine level taken during hospitalisation
        \item Patient glomerular filtration rate in comparison with reference, taken during hospitalisation
        \item Patient maximum C-reactive protein B level taken during hospitalisation
        \item Comorbidity: heart failure during 5 years prior to admission \textit{[Y/N]}
        \item Comorbidity: diabetes during 5 years prior to admission \textit{[Y/N]}
        \item Comorbidity: cancer during 5 years prior to admission \textit{[Y/N]}
    \end{itemize}
    
    \item Using these attributes, and given two hyper-parameters $C$ and $P$ that determine, respectively, minimum cardinality of each group and of each subset within each group containing patients labelled with only one of \textit{[Y/N]}, patients belonging to the \textit{training data set} are initially stratified in $m$ groups ${\cal G}_1, ..., {\cal G}_m$ using a standard similarity-based clustering algorithm, e.g. \textit{k-means clustering} \cite{ostrovsky2013effectiveness}, in such way that $m$ is the maximal number of groups satisfying the constraints imposed by $C$ and $P$. These constraints are imposed in order to ensure that each group is sufficiently large and sufficiently diverse to generate effective predictors.
    
    In our experiments, we have set up the hyper-parameters $C$ and $P$ respectively as $200$ and $50$. With these values, the final number of groups is $m = 3$.
    
    \item A heuristic predictor, based on a specified \textit{supervised learning} technique, is built for each group ${\cal G}_i$. 
    
    In our experiments, we have employed \textit{GAM} predictors, which are flexible enough as to build nonlinear hyper-surfaces separating patients with different labels, and intuitively self-explanatory enough to build classification criteria which are meaningful to domain experts.
    
    \item Each group ${\cal G}_i$ is sub-divided according to values of the decision labelling attribute, this way forming \textit{poles} -- in our case, ${\cal G}_i^Y$ and ${\cal G}_i^N$ corresponding to mortality prior to 90 days post-discharge \textit{[Y/N]}.
    
    \item The \textit{centroid} of each pole is defined as a tuple of values of patient attributes (excluding decision labelling attributes), in which each value is the mean value of that attribute considering all patients belonging the pole.
    
    \item Each patient $\cal P$ in the \textit{validation data set} is assigned to a group, according to the following heuristics: $\cal P$ is allocated to group ${\cal G}_i$ if the Euclidean distance between the attribute values of $\cal P$ (excluding the decision labelling attributes) and of one of the centroid poles ${\cal G}_i^Y$ and ${\cal G}_i^N$ is minimal considering all poles of all centroids.
    
    \item Using the corresponding predictor for each patient in the \textit{validation data set}, each patient is classified with respect to the probability of having the decision labelling attribute value as \textit{Y}.
    
    \textit{AUROC} is empirically estimated for the \textit{validation data set} considering the actual value of the decision labelling attribute and the estimated probability for each patient in that data set. The sum of \textit{AUROC} of all groups is defined as the heuristic assessment of quality of the groups.
    \item Given the hyper-parameter $b$ that determines the number of patients to be reallocated across groups at each step during traversal of the search space, two groups are randomly selected and identified respectively as \textit{source} and \textit{target} groups. $b$ patients are randomly selected from the \textit{source} group and tentatively reallocated to the \textit{target} group.
    
    In our experiment, we have set up the hyper-parameter $b$ as $50$, corresponding to $\approx 0.5\%$ of the \textit{training data set}.
    \item Steps 4-9 are repeated with the updated groups of patients. If the sum of \textit{AUROC} of all groups for the \textit{validation data set} obtained using the updated heuristic predictors is larger than the previously obtained value, then the updated groups are kept, otherwise they are discarded.
    \item Given the hyper-parameter $N$, steps 4-10 are repeated $N$ times. 
    
    In our experiments, we have set up $N$ as 5. 
    
    \item The final predictors are then applied to the \textit{test data set}, in order to build sufficiently reliable estimates for the accuracy and robustness of the obtained predictors.
\end{enumerate}

\begin{figure}
    \centering
    \includegraphics[scale=0.45]{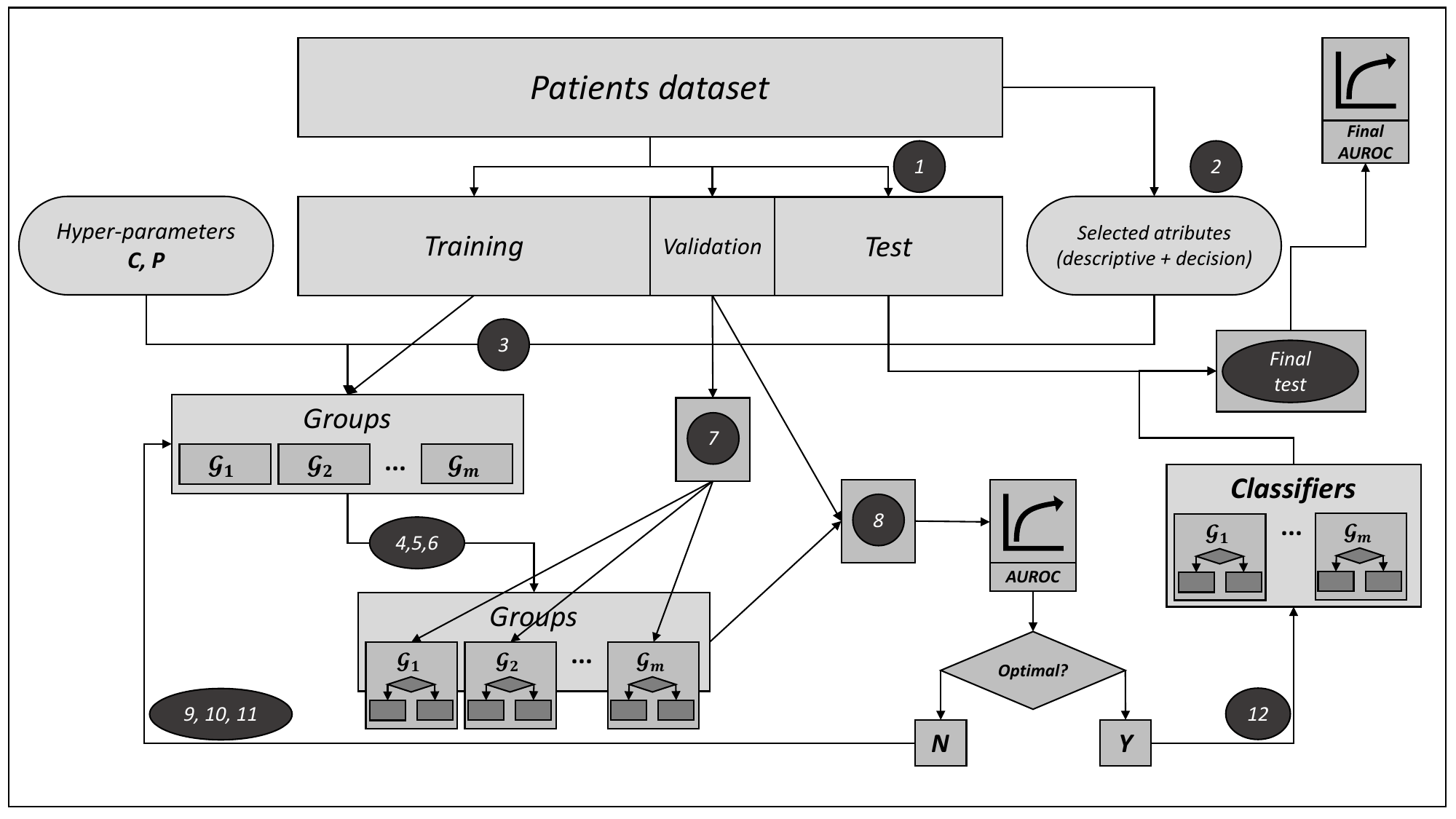}
    \caption{Implementation execution workflow}
    \label{fig:implementation_pdf}
\end{figure}

A diagram depicting the workflow of this implementation is presented in Figure \ref{fig:implementation_pdf}. In this diagram, each step number is presented inside circles and can be followed along with the steps described in the previous paragraphs.

This procedure is asymptotically convergent, and the resulting groups of patients have been empirically assessed as efficient for the construction of accurate prediction models for the decision labelling attribute, the same way as occurs with the original \textit{Clustering Aware Classification} model \cite{srivastava2023clustering}.

The hyper-parameters $C, P, b$ and $N$ must be determined empirically and based on expert knowledge. Experimental results indicate that the values selected for the hyper-parameters are sufficient to build convergence towards groups of sufficiently similar patients. The construction of groups based on maximisation of \textit{AUROC} and allocation of patients to groups based on similarity with pole centroids are grounded on heuristic assumptions that the actual population of patients under consideration contains diverse sub-groups for which specialised predictors should be employed. The selection of \textit{GAM} predictors is grounded on the assumption that the classes to be identified by each specialised predictor admit a separator that can be formally characterised as a hyper-plane or as a smooth differentiable manifold (intuitively, a smooth surface that is not necessarily linear). Linear separators such as logistic regression or Fisher linear discriminant \cite{james2013introduction} are robust, self-explanatory and computationally efficient, hence they should be welcome when they prove to fit well with particular problems and produce accurate and reliable predictions. Separators such as \textit{GAM} preserve these features while permitting the separation of more complex classes, by replacing separating hyper-planes with separating smooth, differentiable manifolds. Intuitively, it means that complex, nonlinear separators can be effectively replaced by a collection of separators according to groups of patients built according to the heuristic method proposed here.

The information that is available for predictions is imperfect in many senses:
\begin{itemize}
    \item The population of patients used to build models is restricted and has imprecise boundaries, in the sense that new patients are intuitively expected to be in or out of the population according to clinical assessment. Hence, decision whether prediction outcomes apply to a new patient are not fully formalised.
    \item The selected attributes to characterise each patient, and their corresponding characterisation based on measurement units, resolution etc. depends on expert knowledge. The quality of predictions is limited by how accurately new patients can be portrayed through the selected attributes.
    \item Assumptions about how attributes relate to the selected decision labelling attribute indicate a family of prediction models that is assumed to be appropriate for the problem under consideration. These assumptions can only be partially formalised and verified.
    \item Relations involving attributes are based on causal links which are assumed \textit{a priori}, some of which can be tested using statistical methods. Unconsidered relations cannot be easily detected, nor can be identified relations involving attributes which -- even unknown -- can be influential in the generation of observable co-occurrences of values. Interpretations of co-occurrences as significant correlations or causal links can, at best, be assumed and tested.
\end{itemize}

Step 12 -- the final step in the allocation procedure -- is included as a means to infer the behaviour of the predictors when facing data of new patients. The accuracy and robustness of the predictors is estimated by the empirical \textit{AUROC} obtained for each group of patients in the \textit{test data set}, as well as the \textit{empirical error rate} observed in each group and an \textit{estimated upper bound for the error rate} based on \textit{Agnostic PAC Learning} assessment \cite{kalai2012reliable,shalev2014understanding}.

\textit{Probably Approximately Correct (PAC) machine learning} refers to a rigorous characterisation of learning \textit{problems} -- particularly supervised learning problems, in which prediction models are based on identification of co-occurrence patterns involving general patient attributes and one decision labelling attribute -- in such way that \textit{sample size, reliability of predictions} and \textit{accuracy of predictions} can be connected: under clearly specified conditions that characterise, in terms of statistical regularities, the context within which the prediction models are expected to work, we have that the sample size expressed as a minimal number of items $\omega$, the accuracy of predictions expressed as a maximal error rate $\epsilon$, and the reliability of predictions expressed as a maximal probability $\delta$ of observation of prediction errors above a certain threshold, can be connected in equational results indicating that 

\begin{center}
    $\epsilon = {\cal F}(\delta, \omega)$.
\end{center}

In other words, given $\delta$ and $\omega$, we can deduce relations indicating that, given a sufficiently large sample of size at least $\omega$, with probability at least $(1 - \delta)$ the error rate in predictions is bounded from above by $\epsilon$ \cite{valiant1984theory}.

\textit{Agnostic PAC learning} extends the original theory of \textit{PAC learning} by relaxing the conditions upon which the equational results can be grounded.

In our specific case, based on the results upon which \textit{Clustering Aware Classification} is founded \cite{srivastava2023clustering}, we have:

\begin{itemize}
    \item Given a number $\omega$ of post-AKI patients in the \textit{test data set}, for whom all attributes considered relevant to predict 90 days post-discharge mortality risk are available, including a decision attribute ``has patient died within 90 days after discharge? \textit{[Y/N]}", a risk measure $P_i$ for each patient $i$ is predicted -- in the form of the probability of mortality within the prescribed period after discharge -- and confronted with the decision attribute the following way: if the decision attribute label for the specific patient $i$ is $Y$, then the error is computed as $er_i = (1 - P_i)$, otherwise it is computed as $er_i = P_i$.
    \item The \textit{empirical error} is calculated as:
    \begin{center}
        ${\cal L}^{emp} = \frac{\sum_i^\omega er_i}{\omega}$.
    \end{center}
    \item The \textit{error bound based on empirical Rademacher complexity} indicates uncertainty about error estimates due to bias in prediction models. In the case of linear separators, it is parameterised by the $L_2$  norm of the linear weights of the separator $||\vec{w}||$ and can be calculated as:
    \begin{center}
        ${\cal R}^{emp} = 2 \sqrt{\frac{||\vec{w}||}{\omega}}$
    \end{center}
    \item The \textit{reliability based error bound} indicates uncertainty about error estimates due to sample size. In the case of linear separators, it is parameterised by the reliability parameter $\delta$ and can be calculated as:
    \begin{center}
        ${\cal U} = \sqrt{\frac{ln(1/\delta)}{2 \omega}}$
    \end{center}
    
    Since \textit{GAM} models build quasi-linear separators, we employ the same measures of Rademacher complexity  and reliability here.
    \item Given all these factors, the \textit{overall error upper bound} can be estimated as:
    \begin{center}
        ${\cal L} = {\cal L}^{emp} + {\cal R}^{emp} + {\cal U}$
    \end{center}
\end{itemize}

In the following section we present experimental results using these heuristics.

 \section{Results and discussion}
\label{sec:results}

Our experimental results are based on two data sets:

\begin{enumerate}
    \item A synthetic data set, custom built to explore how the heuristics work; and
    \item A data set extracted from the \textit{Grampian Data Safe Haven}, which is a large repository belonging to a network of observational data sets comprising public and population health information in Scotland.
\end{enumerate}

The synthetic data set was built the following way:

\begin{itemize}
    \item We consider an abstract population of $1500$ data items, and the attributes $X_1, X_2, X_3, X_4$ and $Y$, such that:
    \begin{itemize}
        \item $X_3 = X_1 + X_2$ and $X_4 = X_1 - X_2$. We assume that we do not have direct access to $X_1, X_2$, only to their functionally dependent variables $X_3, X_4$.
        \item $Y$ is the decision labelling attribute.
    \end{itemize}
    \item In one half of the data items, the values of the attributes are defined as follows:
    \begin{itemize}
        \item $X_1$: uniformly distributed real values in $[-0.8, 0.2]$.
        \item $X_2$: uniformly distributed real values in $[0, 1]$.
        \item $Y$: $\left\{ \begin{tabular}{l}
                            if $X_1 + X_2 \leq 0$: \textit{Y} \\
                            otherwise: \textit{N}
                            \end{tabular}
                    \right.$
    \end{itemize}
    \item In the other half of the data items, the values of the attributes are defined as follows:
    \begin{itemize}
        \item $X_1$: uniformly distributed real values in $[0, 1]$.
        \item $X_2$: uniformly distributed real values in $[-0.8, 0.2]$.
        \item $Y$: $\left\{ \begin{tabular}{l}
                            if $X_1 + X_2 > 0$: \textit{Y} \\
                            otherwise: \textit{N}
                            \end{tabular}
                    \right.$
    \end{itemize}
 \end{itemize}
 
 This synthetic data set is crafted in such way that, without separation of data items in groups, any hyper-surface that separates the decision values is less effective than the specialised hyper-surfaces based on inferred groups. After separation in two groups, prediction models feature higher accuracy and reliability.
 
We have manually adjusted the values of the hyper-parameters to build two groups, and readjusted block size $b$ to 1 and the number of rounds $N$ to 10. With these values, we obtained the results featured in Table \ref{table:tab1}. Columns ${\cal L}^{emp}, {\cal L}$ and \textit{AUROC} correspond respectively to the \textit{empirical error}, the estimated \textit{error upper bound} and the \textit{AUROC} with corresponding interval limits with respect to $95\%$ confidence. Rows correspond to the two inferred groups ${\cal G}_1$ and ${\cal G}_2$, to the overall data set (\textit{ALL}), and to the utilisation of plain logistic regression (hence, a linear separator) as predictor on the overall data set (\textit{ALL-logit}). Estimates of the \textit{error upper bounds} are relatively large -- and larger for group based predictors than for the case of the generic predictor -- mostly due to sample sizes, as we split the data set into 400 patients for \textit{training}, 400 other patients for \textit{validation} and 700 patients for \textit{testing}. Our main goal with this experiment is to illustrate that the heuristics are able to identify the two groups accordingly, even without using the variables $X_1, X_2$ directly and in the scenario in which -- by design -- there is overlap between feature values that determine the use of different decision rules.

In our experiments, only 4 of the 700 patients in the \textit{test data set} were misclassified upon utilisation of group based predictors. As can be seen in Table \ref{table:tab1}, the empirical performance of specialised predictors is higher than that of the generic predictor that was built using the whole \textit{training data set}, and even higher than that of logistic regression. Estimated overall performance of specialised predictors is low, given that the synthetic data contains bias and, as a consequence, features artificially large empirical Rademacher complexity.

\begin{figure}[htbp]
      \centering
      \includegraphics[scale=0.9]{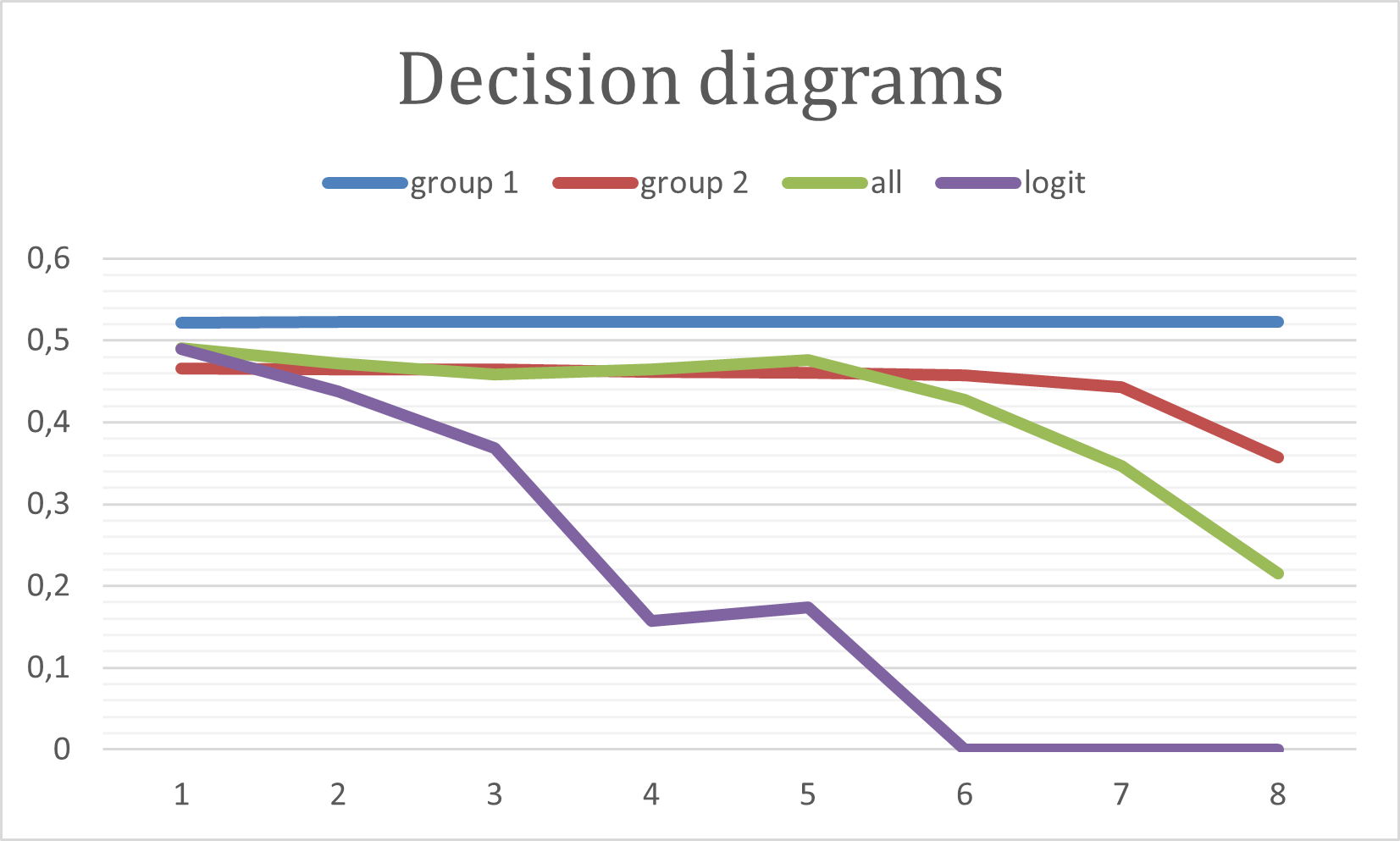}
      \caption{Decision diagrams -- synthetic data set}
      \label{fig:dec_synth}
  \end{figure}

Net benefit decision diagrams for each of group based predictors, \textit{GAM} for overall data set and logistic regression for overall data set are depicted in Figure \ref{fig:dec_synth}. Improved stability of group based predictors over predictors based on overall data set can be visually verified through these diagrams. The eight points in the X-axis correspond resp. to the decision thresholds \textit{0.01, 0.1, 0.2, 0.4, 0.5, 0.6, 0.8, 0.95}.
 
\begin{table}[htbp]
    \centering
    \begin{tabular}{|c|c|c|c|}
    \hline
                    & ${\cal L}^{emp}$ &  ${\cal L}$ & \textit{AUROC} \\
    \hline
    \hline
     ${\cal G}_1$ & $2 \times 10^{-6}$     &  $1$     & $1 [1, 1]$\\
    \hline
     ${\cal G}_2$ & $6 \times 10^{-3}$     &  $1$     & $1 [1, 1]$\\
    \hline
     \textit{ALL} & $0.04$ & $0.1$ & $0.998 [0.997, 0.999]$\\
    \hline
     \textit{ALL-logit} & $0.25$ & $0.32$     & $0.53 [0.48, 0.58]$\\
    \hline
    \end{tabular}
    \caption{Empirical and estimated limits for errors, and \textit{AUROC} by group and for whole data set -- synthetic data set}
    \label{table:tab1}
\end{table}

We have also tested this approach with real data extracted from the \textit{Grampian Data Safe Haven (G-DaSH)}, managed by the University of Aberdeen, UK, which is part of the Scottish network of Data Safe Havens, a federation of carefully managed and safeguarded patient data repositories available for research \cite{gaggioli2019positive}.

We find in the G-DaSH a collection of around 20 thousand patients with clinical histories including AKI and hospitalisation\footnote{Individualised patient data are obfuscated in the present article to safeguard anonymity of patients in G-DaSH.}. These patients were split into a \textit{training data set} of around 50\% of the data set, a \textit{validation data set} of around 10\% of the data set thousand patients, and a \textit{test data set} of around 40\% of the data set.

We then followed the steps indicated in the previous section, to establish the following:

\begin{enumerate}
    \item We selected a collection of attributes, including demographic attributes (age, sex), clinical measures (fluctuations in creatinine level, estimated glomerular filtration rate, estimated C reactive protein) and relevant comorbidities (heart failure, diabetes, cancer).
    \item We followed the heuristics described in the previous section to stratify the patients using the \textit{training data set} and the \textit{validation data set} in \textit{three groups} ${\cal G}_1, {\cal G}_2, {\cal G}_3$ in such way as to optmise prediction of 90 days post discharge mortality risk.
    \item We built \textit{GAM} predictors for each group. As additional tests, we also built a \textit{GAM} predictor and a linear predictor based on logistic regression for the \textit{training data set} as a whole.
    \item Using the proposed heuristics, we then stratified the patients in the \textit{test data set} in the obtained groups.
    \item We employed the appropriate prediction model for each group in the \textit{test data set} and assessed the accuracy of each model. We have also assessed the accuracy of the models built using the \textit{training data set} as a whole.
    \item Finally, we looked at the groups which were built using the \textit{training data set}, in order to identify patterns in attributes that could explain the classification results obtained for each group.
\end{enumerate}

The obtained results are as follows:

\subsection{Empirical error, error upper bound and \textit{AUROC}}

The \textit{empirical error} ${\cal L}^{emp}$, \textit{estimated error} ${\cal L}$ and \textit{AUROC} with corresponding $95\%$ confidence interval limits for each group ${\cal G}_i, i = 1, ..., 3$, as well as for the models built using the whole \textit{training data set} and applied to the whole \textit{test data set} using \textit{GAM} and logistic regression, are presented in Table \ref{table:tab2}. Additionally, the approximate number of patients belonging to each group is presented ($m$).

\begin{table}[htbp]
    \centering
    \begin{tabular}{|c|c|c|c|c|}
    \hline
                    & ${\cal L}^{emp}$ &  ${\cal L}$ & \textit{AUROC} & $m$\\
    \hline
    \hline
     ${\cal G}_1$ & $0.11$ &  $0.27$ & $0.77 [0.72, 0.83]$ & $550$\\
    \hline
     ${\cal G}_2$ & $0.10$ &  $0.16$ & $0.74 [0.72, 0.77]$ & $2650$\\
    \hline
     ${\cal G}_3$ & $0.09$ &  $0.14$ & $0.71 [0.69, 0.73]$ & $5000$\\
    \hline
     \textit{ALL} & $0.10$ &  $0.13$ & $0.73 [0.71, 0.75]$ & $8200$\\
    \hline
     \textit{ALL-logit} & $0.10$ &  $0.13$ & $0.72 [0.70, 0.73]$ & $8200$\\
    \hline
    \end{tabular}
    \caption{Empirical and estimated limits for errors, and \textit{AUROC} by group and for whole data set -- G-DaSH}
    \label{table:tab2}
\end{table}



As expected, larger data sets are more robust than smaller ones, hence estimated errors are smaller in the experiments based on G-DaSH in comparison with those obtained using the much smaller synthetic data set. Interestingly, however, \textit{AUROC} does not increase with the size of data sets.

No significant differences in accuracy or \textit{AUROC} were observed when using the group based predictors, in comparison with the generic models when using observational data. As described in the next subsection, these empirical results suggest that the main value of the stratification of patients in groups lies not on improving accuracy in predictions, but on unveiling groups of patients featuring diversified profiles with respect to how attributes connect to decision making variables.

\begin{figure}[htbp]
      \centering
      \includegraphics[scale=0.9]{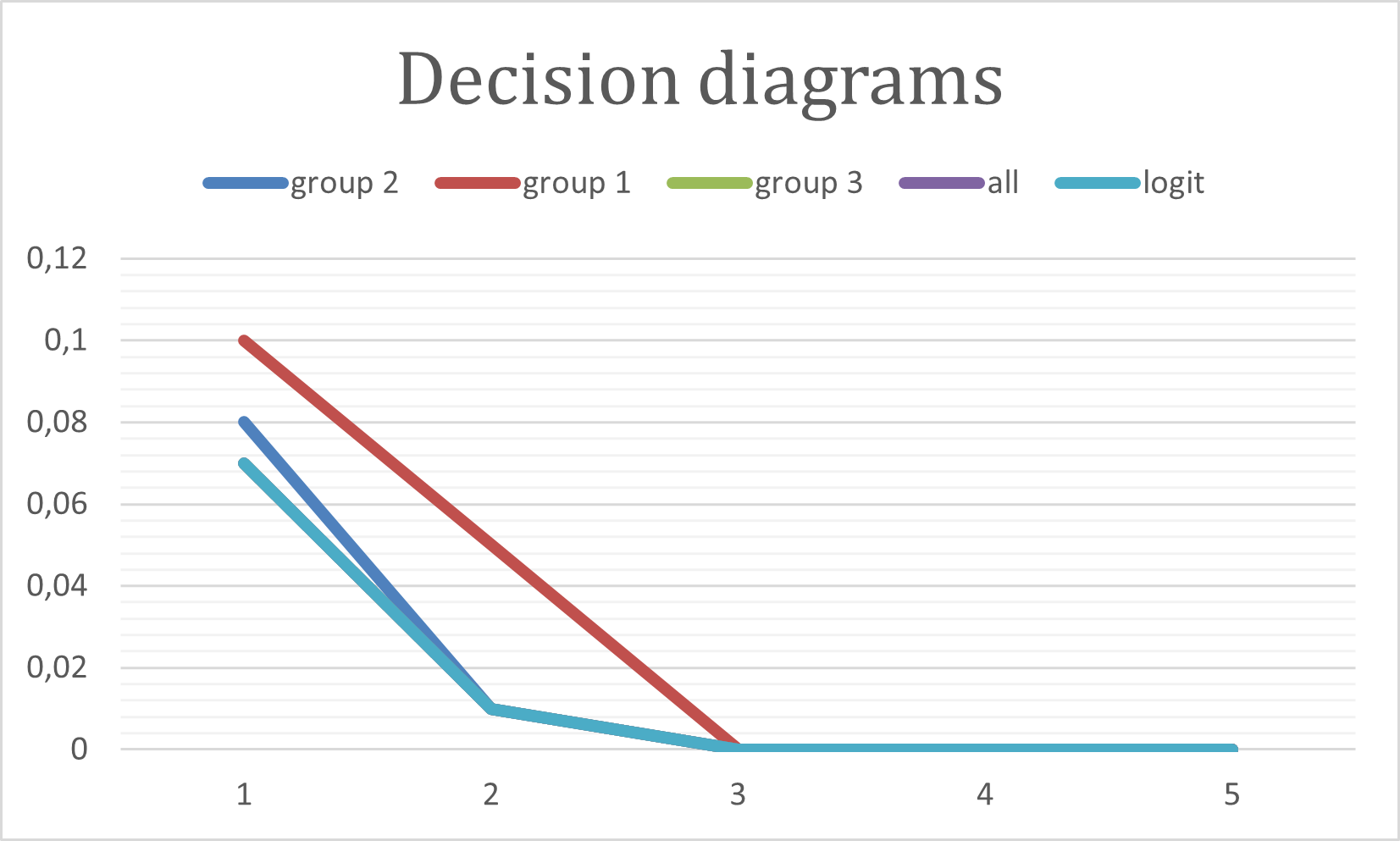}
      \caption{Decision diagrams -- G-DaSH data set}
      \label{fig:dec_dash}
  \end{figure}

Net benefit decision diagrams for each of group based predictors, \textit{GAM} for overall data set and logistic regression for overall data set using G-DaSH data are depicted in Figure \ref{fig:dec_dash}. The five points in the X-axis correspond resp. to the decision thresholds \textit{0.05, 0.2, 0.5, 0.8, 0.95}.

\subsection{Attribute profiling within groups of the training data set}

The experimental results with the synthetic data set as well as with the large observational data set provided by G-DaSH indicate that the heuristics to allocate new patients to appropriate groups are acceptable. Experimental results with synthetic, tailor made data indicate that the use of group based predictors can improve accuracy and robustness of results, and experimental results with realistic, observational data indicate that, at least, accuracy seems to be preserved in real life situations.

\begin{figure}[htbp]
      \centering
      \includegraphics[scale=0.9]{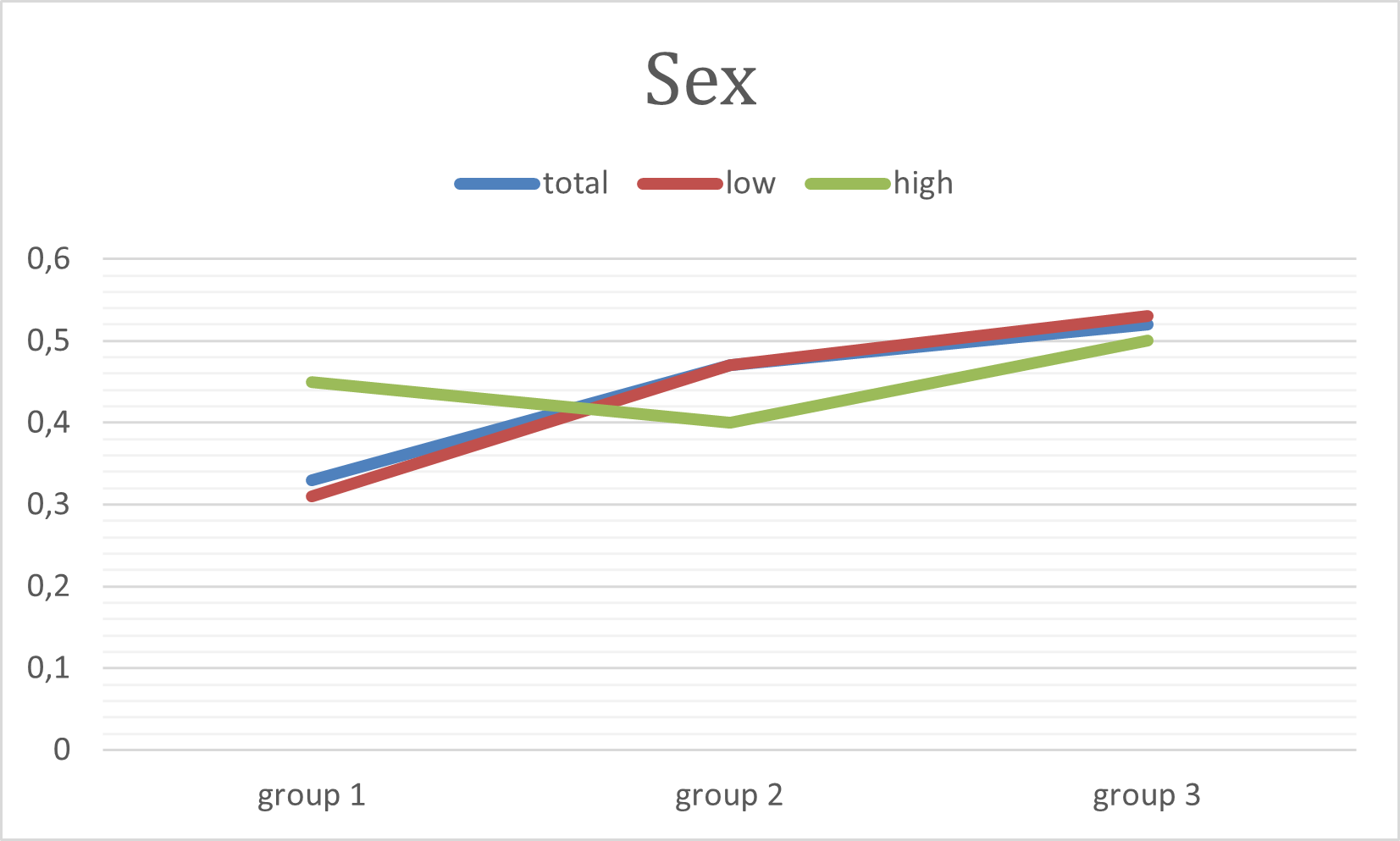}
      \caption{Group based gender distribution}
      \label{fig:sex}
  \end{figure}

\begin{figure}[htbp]
      \centering
      \includegraphics[scale=0.9]{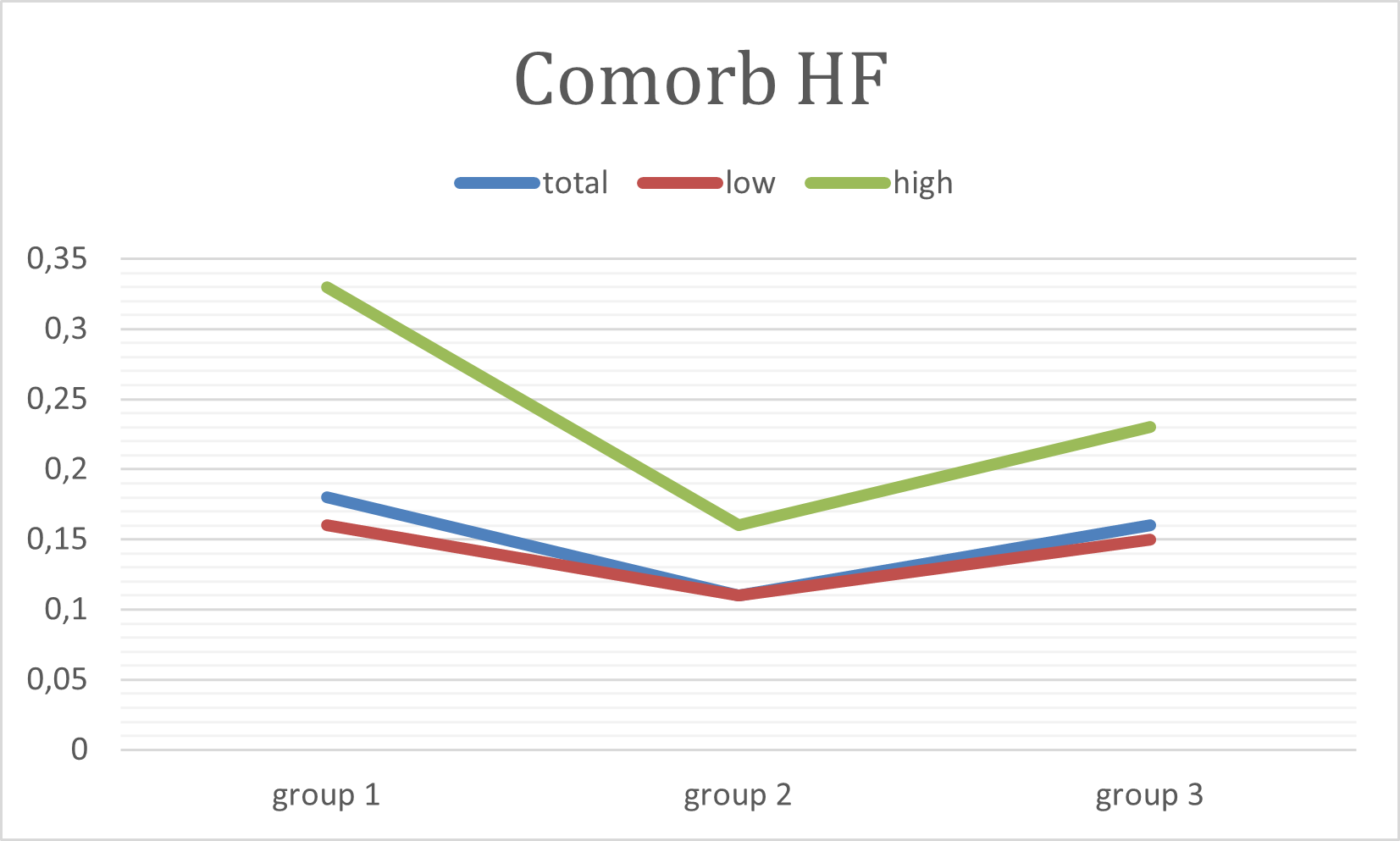}
      \caption{Group based distribution -- history of heart failure}
      \label{fig:HF}
  \end{figure}

\begin{figure}[htbp]
      \centering
      \includegraphics[scale=0.9]{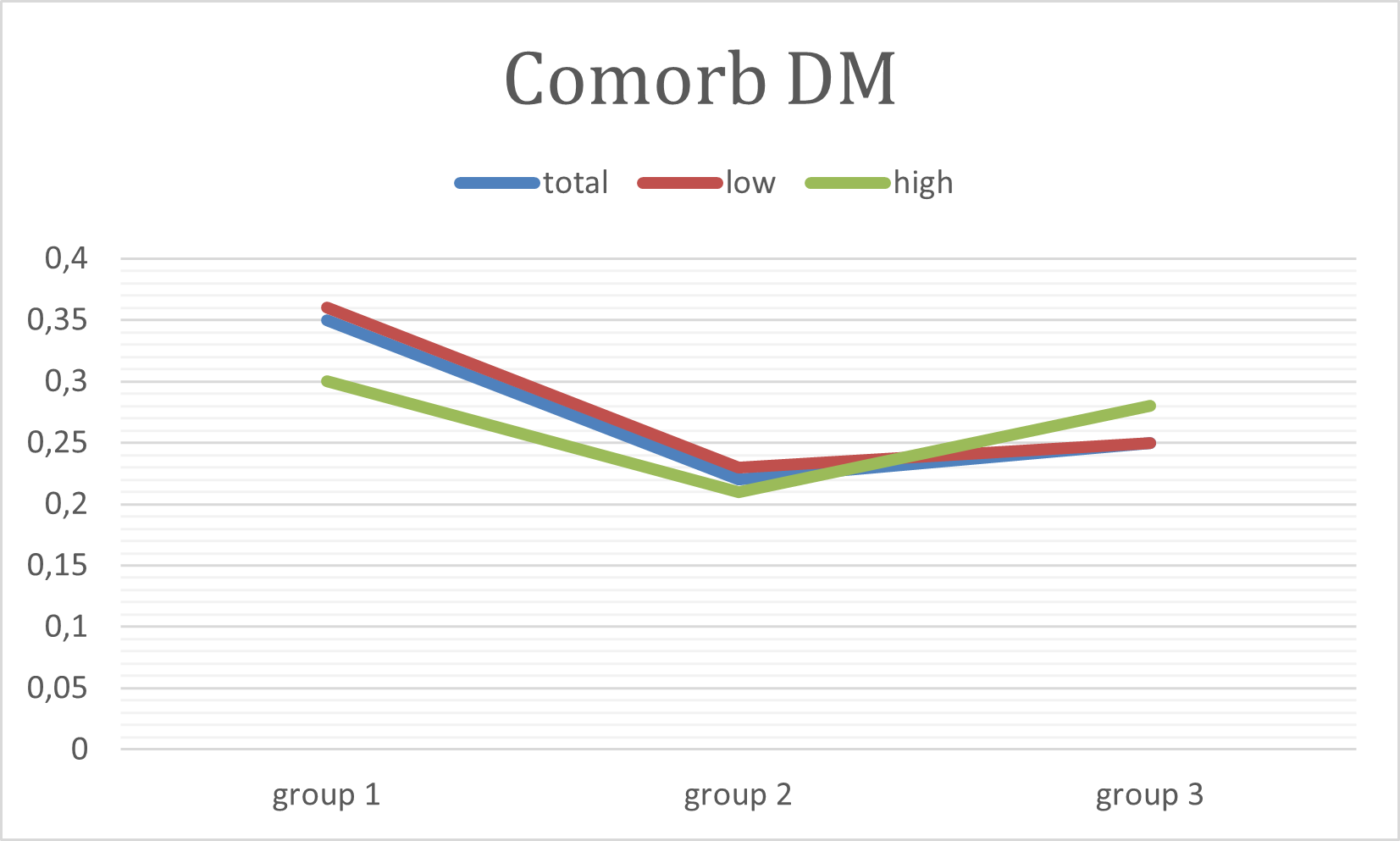}
      \caption{Group based distribution -- history of diabetes}
      \label{fig:DM}
  \end{figure}

\begin{figure}[htbp]
      \centering
      \includegraphics[scale=0.9]{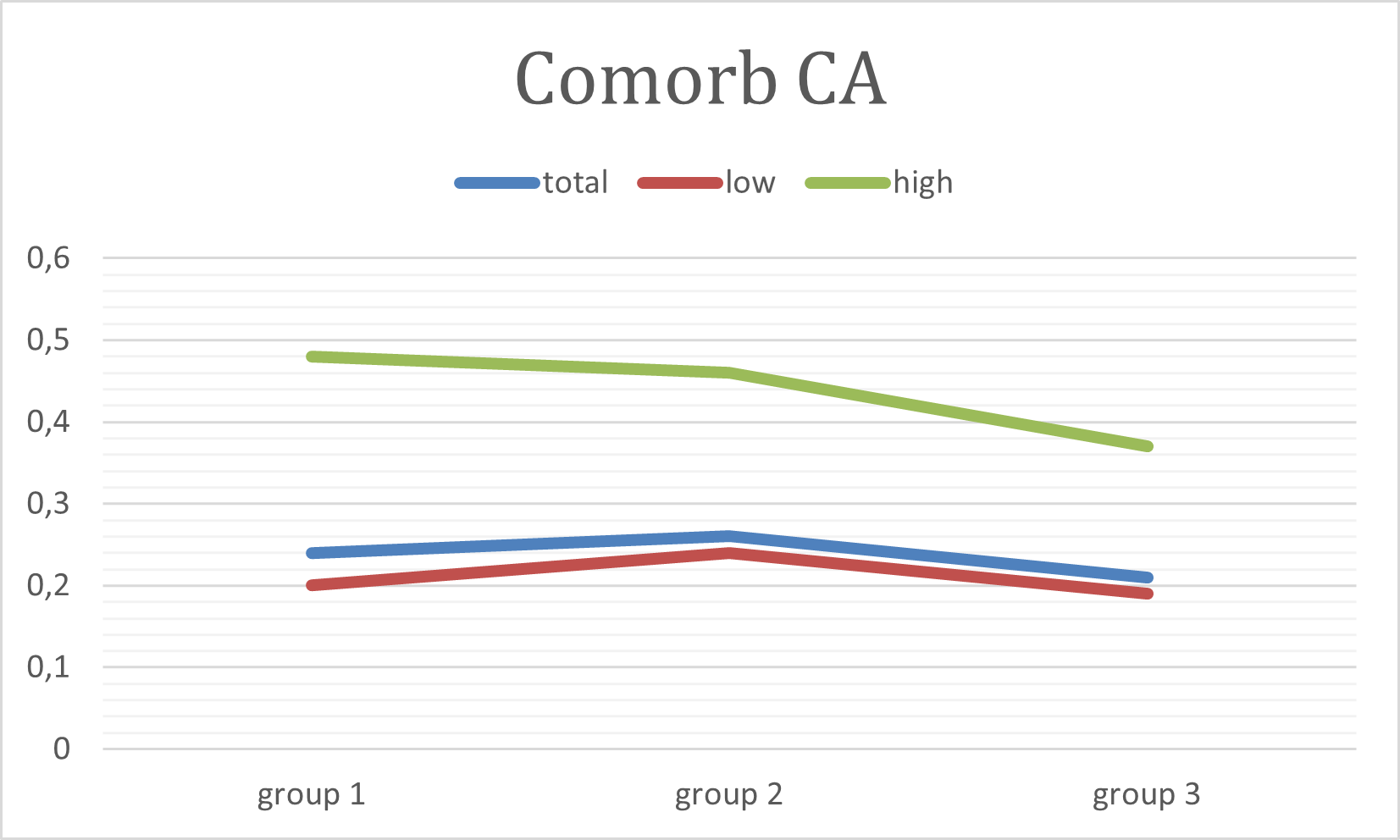}
      \caption{Group based distribution -- history of cancer}
      \label{fig:CA}
  \end{figure}

\begin{figure}[htbp]
      \centering
      \includegraphics[scale=0.9]{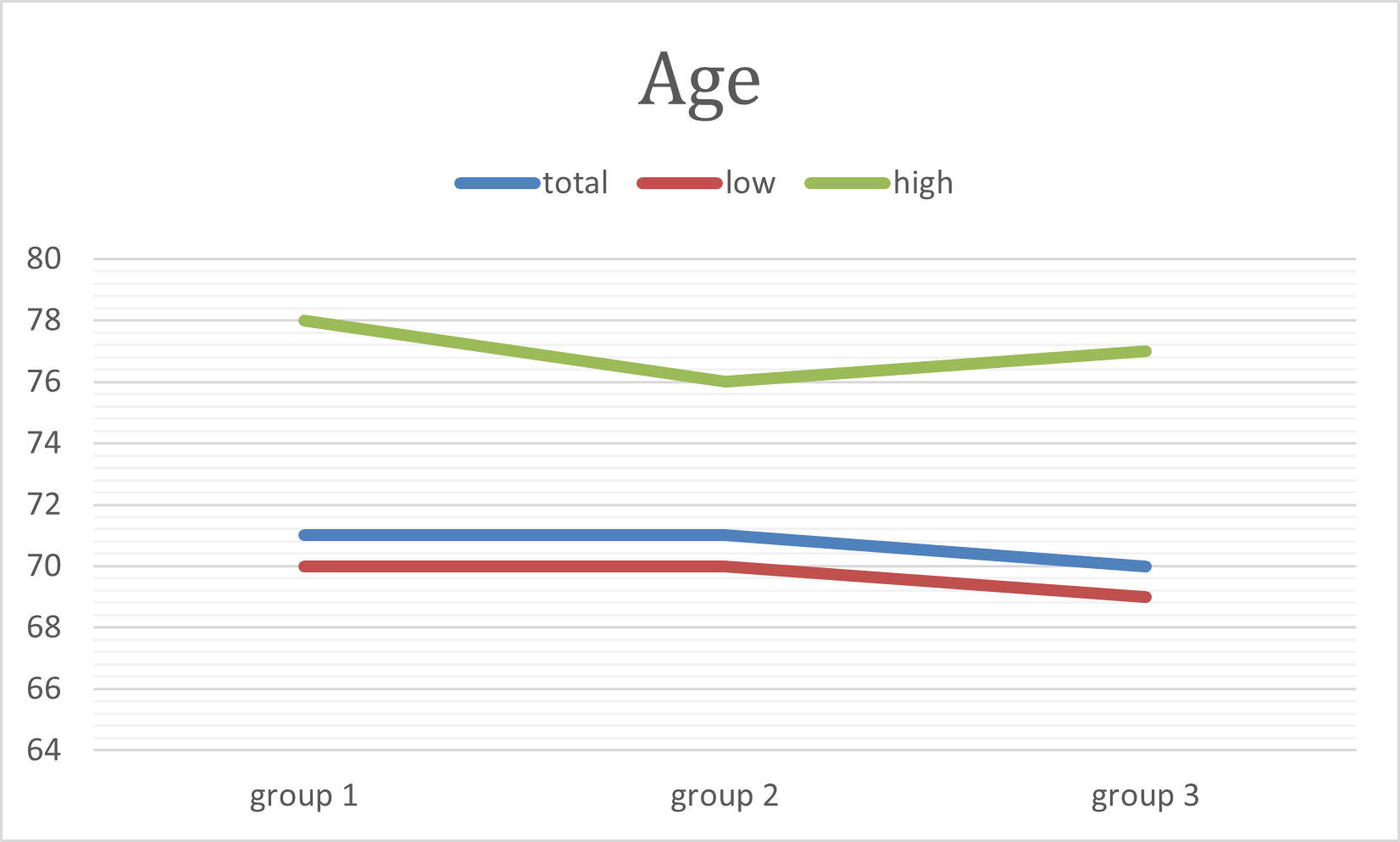}
      \caption{Group based age distribution}
      \label{fig:age}
  \end{figure}

\begin{figure}[htbp]
      \centering
      \includegraphics[scale=0.9]{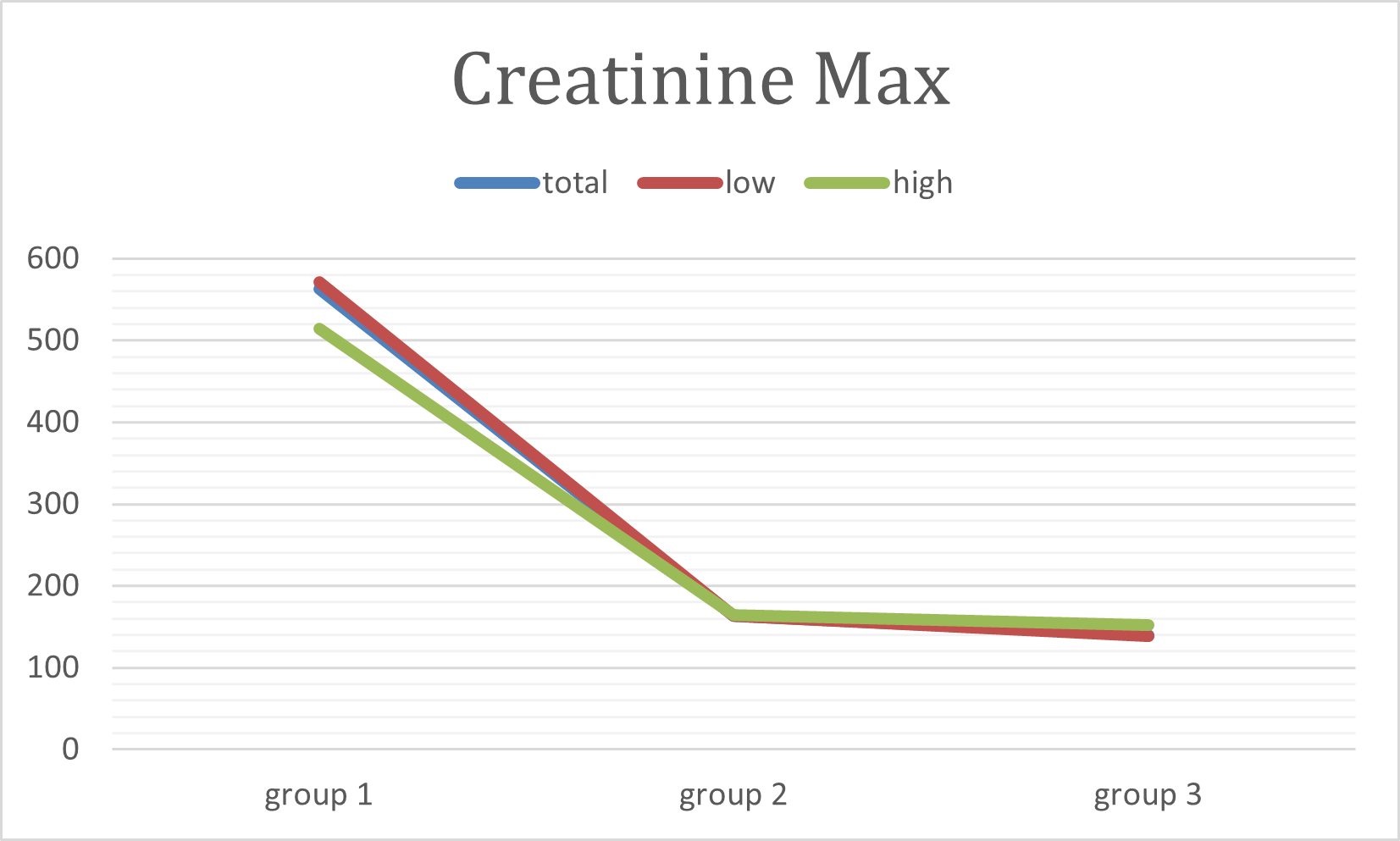}
      \caption{Group based distribution -- maximum creatinine}
      \label{fig:CREAmax}
  \end{figure}

\begin{figure}[htbp]
      \centering
      \includegraphics[scale=0.9]{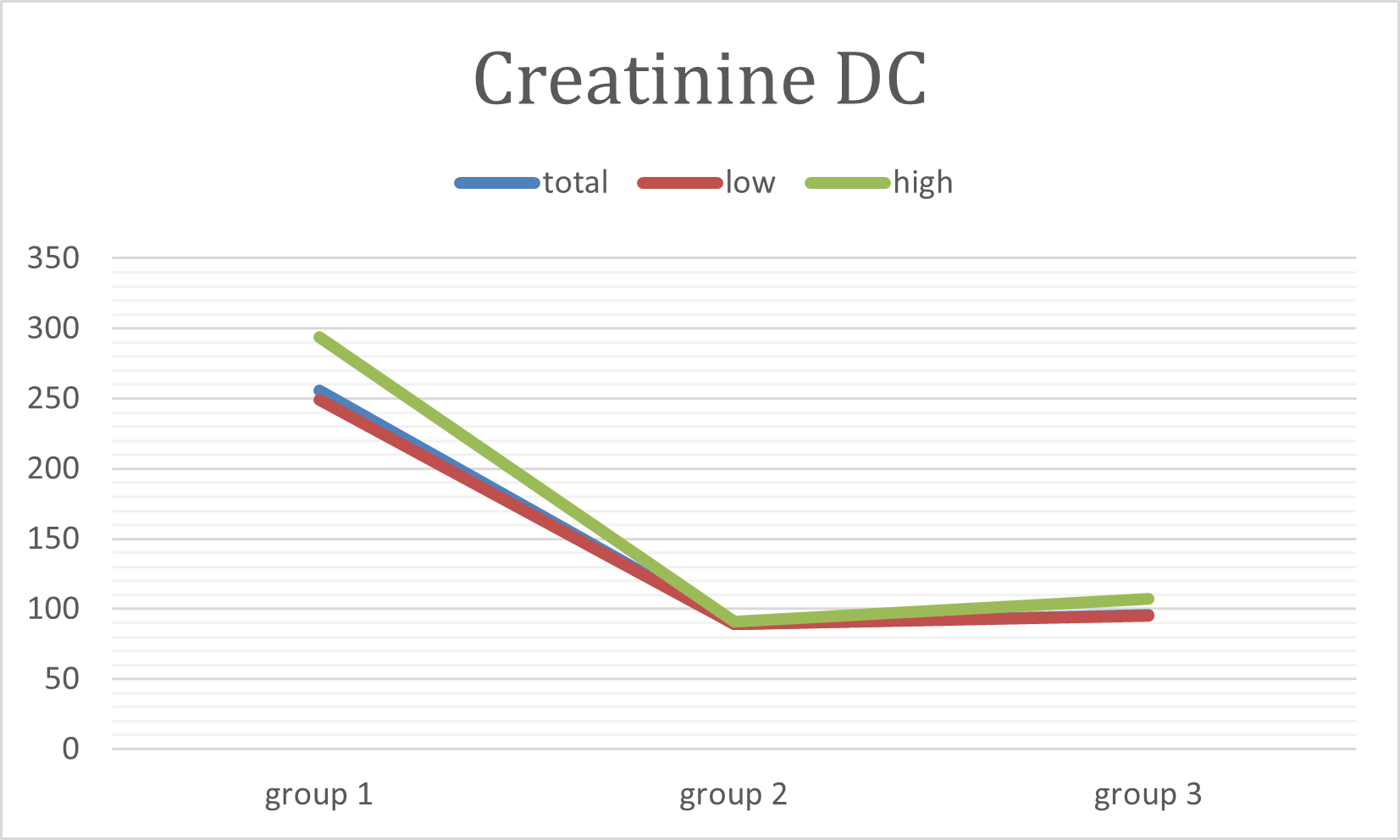}
      \caption{Group based distribution -- creatinine at day of discharge}
      \label{fig:CREAdc}
  \end{figure}

\begin{figure}[htbp]
      \centering
      \includegraphics[scale=0.9]{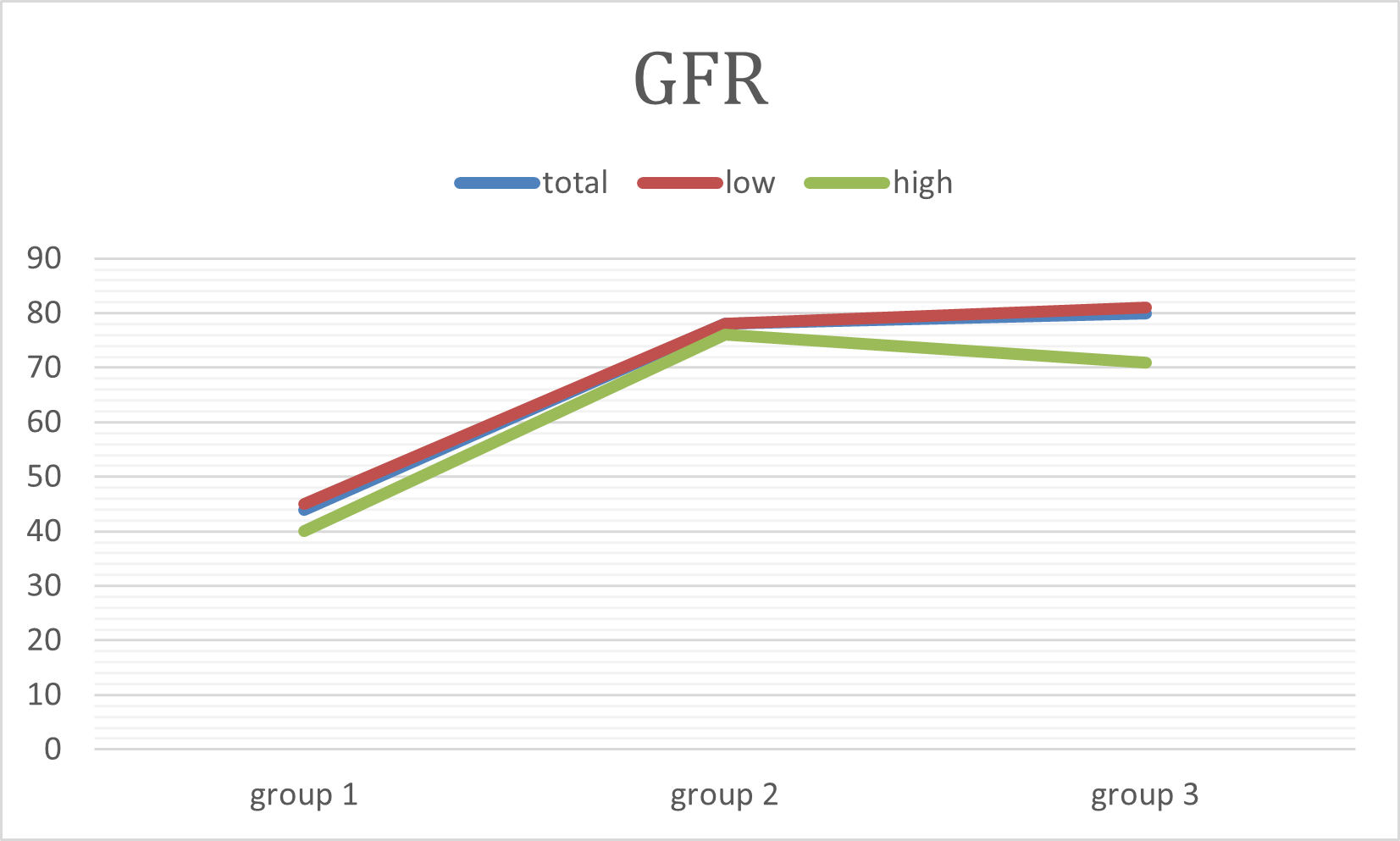}
      \caption{Group based distribution -- GFR}
      \label{fig:gfr}
  \end{figure}

\begin{figure}[htbp]
      \centering
      \includegraphics[scale=0.9]{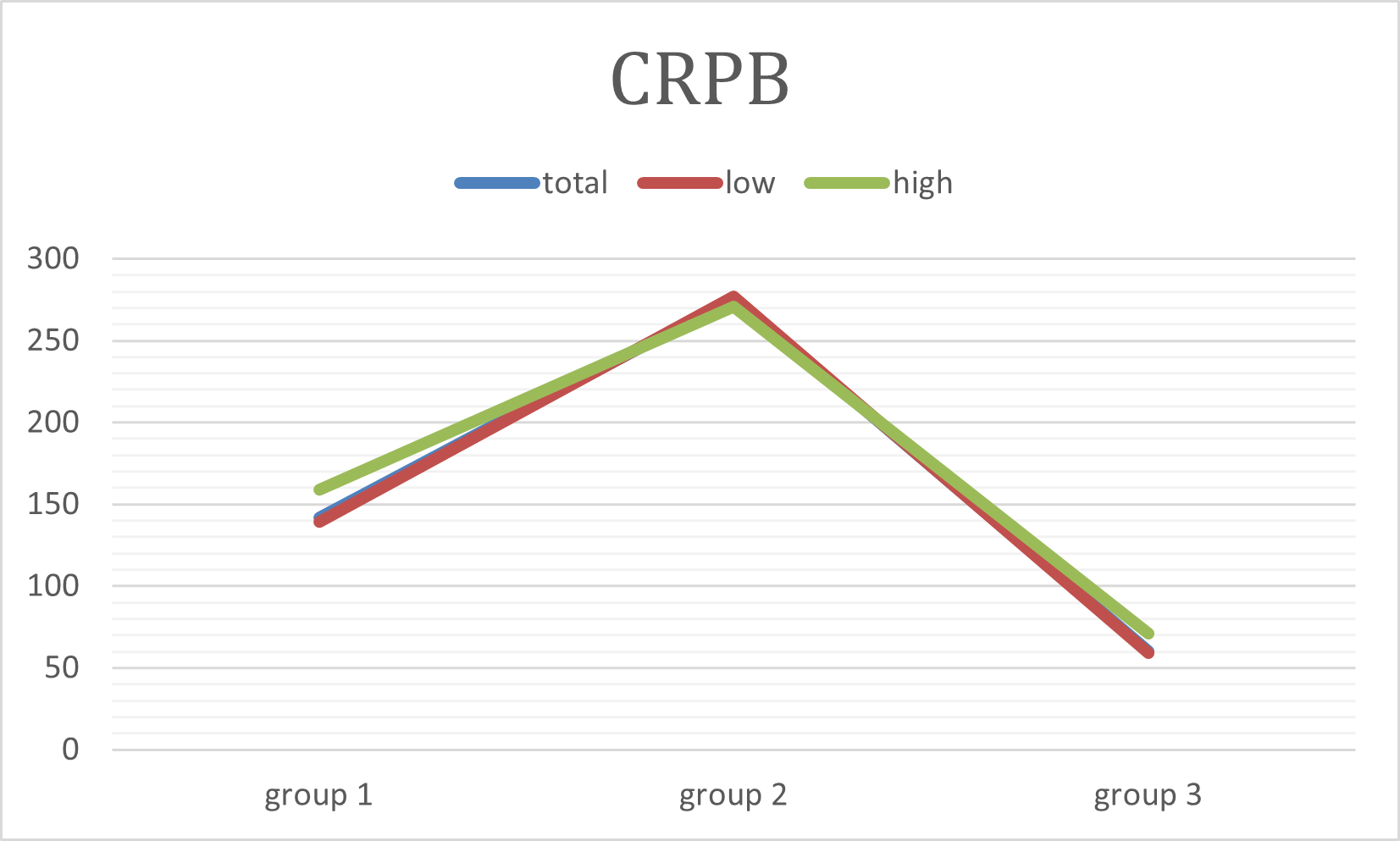}
      \caption{Group based distribution -- CRPB}
      \label{fig:crpb}
  \end{figure}

We have analysed the mean values of the attributes which were used in the experiments to build the \textit{GAM} predictors, considering the actual groups built using the \textit{training data set}. The results of this analysis are as follows:

\begin{itemize}
    \item Gender distribution \textit{as well as the effect of gender in discrimination between low and high mortality risk} were different across groups, as depicted in Figure \ref{fig:sex}.
    \item The distributions of patients with a history of diabetes \textit{as well as the effects of these attributes in discrimination between low and high mortality risk} were different across groups, as depicted in Figure \ref{fig:DM}.
    \item The distributions of patients according to mean values for each of the other attributes were different across groups, but the effects of these attributes in discrimination between low and high mortality risk were consistent across all groups, as depicted in Figures \ref{fig:HF}, \ref{fig:CA}, \ref{fig:age}, \ref{fig:CREAmax}, \ref{fig:CREAdc}, \ref{fig:gfr}, and \ref{fig:crpb}.
\end{itemize}


These experimental results suggest that patients can be stratified in groups with advantages for data generation, given that the interpretation of values to assess mortality risk can change depending on to which group a patient belongs. Therefore, as a general observation, the meaning of values of these attributes changes depending on groups, and the allocation of patients to groups can increase fairness while preserving accuracy.

\section{Conclusions and future work}
\label{sec:conclusion}

The main results from the initiative reported in this article are methodological. We have considered a complex healthcare issue -- namely, mortality risk assessment for post-AKI, post-discharge patients -- and explored the performance of predictions based on population stratification guided by the target prognostics, in order to improve accuracy simultaneously with transparency and fairness, given the identification of groups of patients for whom the interpretation of physiological indicators should be different.

We have discussed how to assess the quality of predictions, considering mathematical and statistical measurements in combination with experience-based considerations provided by an expert clinician. Our empirical results indicate that this strategy is effective for the problem under consideration, indicating that performance of group based predictors outweighs reliability losses due to splitting the initial sample into smaller scale strata. Fairness is nevertheless improved by structuring patients in groups, considering that the interpretation of descriptive variables can be adjusted \textit{per} group.

Finally, we have highlighted throughout the text some fine-grained methodological issues related to the validation and testing of clinical and prognostic strategies. The methodology employed in this work is well aligned with standard protocols used in large scale clinical trials. The obtained results, however, are only indicative of what can be expected from proposed strategies, and should not be misinterpreted as factual estimates of future performance of proposed predictors and treatments given unforeseen patient populations.

One important problem to consider in evidence-based prediction problems that influence therapy strategies is the \textit{drifting target issue}: the utilisation of a system as the one outlined in this work can alter the mortality risks and, therefore, can decrease the accuracy of the system itself. Appropriate strategies to update groups and corresponding predictors until stabilisation is reached should be devised, and this is a relevant next step for the initiative presented in this article. 

A second relevant open problem is the scrutinous clinical characterisation of groups obtained using the heuristics outlined in this work, in order to unveil causes and specialised therapy strategies that should be employed for each group, this way avoiding that minority groups can be treated less effectively than possible because of decisions taken with consideration solely of majority groups.

{
\bibliographystyle{elsarticle-harv}
\bibliography{clinical}
}
\end{document}